\documentclass[10pt,twocolumn,letterpaper]{article}

\usepackage{cvpr}
\usepackage{times}
\usepackage{epsfig}
\usepackage{graphicx}
\usepackage{amsmath}
\usepackage{amssymb}

\usepackage{soul}
\usepackage{url}
\usepackage{mathtools}
\usepackage{bm}
\usepackage{esvect}
\usepackage{enumitem}
\usepackage{subfigure}
\usepackage{multirow}

\newcommand{\cA}{\mathcal{A}}

\newcommand{\cY}{\mathcal{Y}}

\newcommand{\bI}{\mathbf{I}}

\newcommand{\bP}{\mathbf{P}}

\newcommand{\bR}{\mathbf{R}}

\newcommand{\bX}{\mathbf{X}}
\newcommand{\bx}{\mathbf{x}}
\newcommand{\by}{\mathbf{y}}

\newcommand{\bz}{\mathbf{z}}

\newcommand{\cN}{\mathcal{N}}
\newcommand{\bK}{\mathbf{K}}

\DeclareMathOperator*{\argmin}{arg\,min}

\usepackage[pagebackref=true,breaklinks=true,letterpaper=true,colorlinks,bookmarks=false]{hyperref}

\cvprfinalcopy % *** Uncomment this line for the final submission

\def\cvprPaperID{12} % *** Enter the CVPR Paper ID here

\ifcvprfinal\fi

\begin{document}

%%%%%%%%% TITLE
\title{Star Tracking using an Event Camera}

\author{Tat-Jun Chin$^\dagger$ \hspace{1em} Samya Bagchi$^\dagger$ \hspace{1em} Anders Eriksson$^\star$ \hspace{1em} Andr\'{e} van Schaik$^\diamond$\\
The University of Adelaide$^\dagger$ \hspace{1em} University of Queensland$^\star$ \hspace{1em} Western Sydney University$^\diamond$
}

\maketitle
%\thispagestyle{empty}

%%%%%%%%% ABSTRACT
\begin{abstract}
Star trackers are primarily optical devices that are used to estimate the attitude of a spacecraft by recognising and tracking star patterns. Currently, most star trackers use conventional optical sensors. In this application paper, we propose the usage of event sensors for star tracking. There are potentially two benefits of using event sensors for star tracking: lower power consumption and higher operating speeds. Our main contribution is to formulate an algorithmic pipeline for star tracking from event data that includes novel formulations of rotation averaging and bundle adjustment. In addition, we also release with this paper a dataset for star tracking using event cameras\footnote{Visit project website~\cite{site} for the data. This work was supported by AIML and ARC grants LP160100495 and FT170100072.}. With this work, we introduce the problem of star tracking using event cameras to the computer vision community, whose expertise in SLAM and geometric optimisation can be brought to bear on this commercially important application.
\end{abstract}

%%%%%%%%% BODY TEXT
\section{Introduction}
The attitude of a spacecraft is the 3DOF orientation (roll, pitch, yaw) of its body frame with respect to an inertial frame, such as the celestial reference  frame~\cite[Sec.~2.6]{markley14}. Attitude control is a basic functionality in space flight. The subsystem for attitude control is called the Attitude Determination and Control System (ADCS). As the name suggests, there are two main components in an ADCS: estimating the current attitude, and executing a sequence of appropriate signals to the actuators (reaction wheels, thrusters, etc.) to achieve the desired body orientation.

Our work focusses on the attitude determination problem. A number of sensors are in use for estimating spacecraft attitude, such as sun sensors and  magnetometers. It has been established, however, that star trackers are state-of-the-art in spacecraft attitude estimation~\cite{liebe02}, especially to support \emph{high precision} orientation control. As opposed to rough attitude estimation that is used, e.g., in the detumbling process of a satellite, star trackers play a crucial role during stable flight to deliver \emph{fine} attitude estimation to support the mission objectives, e.g., precisely aiming  on-board instruments at a target region in space or on Earth.

\begin{figure}[t]\centering
\includegraphics[width=0.7\columnwidth]{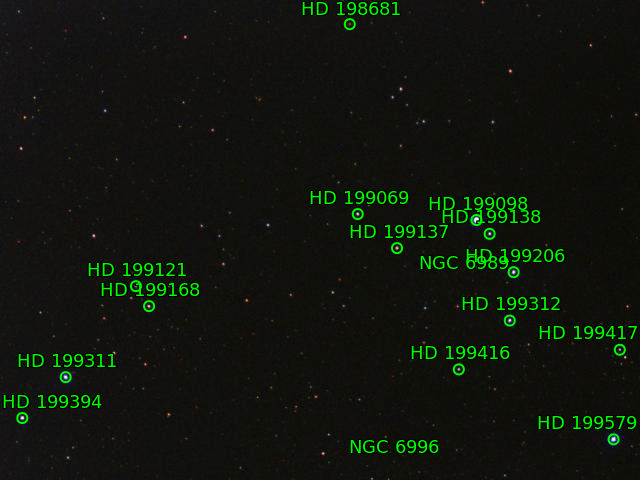}
\caption{Star identification result using the method of~\cite{lang10,astrometry}. The labels correspond to identified stars from a star catalogue.}
\label{fig:astrometry}
\end{figure}

A star tracker is essentially a camera with an image processing algorithm to estimate spacecraft attitude by recognising star patterns~\cite[Chap.~4]{markley14}. Underpinning star tracking is the ability to perform \emph{star identification}~\cite{spratling09} from an image; see Fig.~\ref{fig:astrometry}. In computer vision terms, this means to extract a set of 2D-3D correspondences $\{ (\bx_p, \bX_p  ) \}^{N}_{p=1}$ between the input image and a star chart, where $\bx_p \in \mathbb{R}^2$ are the 2D coordinates of an observed star in the image, and $\bX_p \in \mathbb{R}^3$ is a unit vector that represents the direction of the same star in the inertial frame\footnote{Since stars are effectively at infinity, only their directions matter.}. In fact, the matching is often accomplished by comparing local descriptors (e.g., the geometric hash code of~\cite{lang10}) that encode the spatial configuration of stars in local image regions.

\begin{figure*}[t]\centering
\subfigure[]{\includegraphics[width=0.25\textwidth]{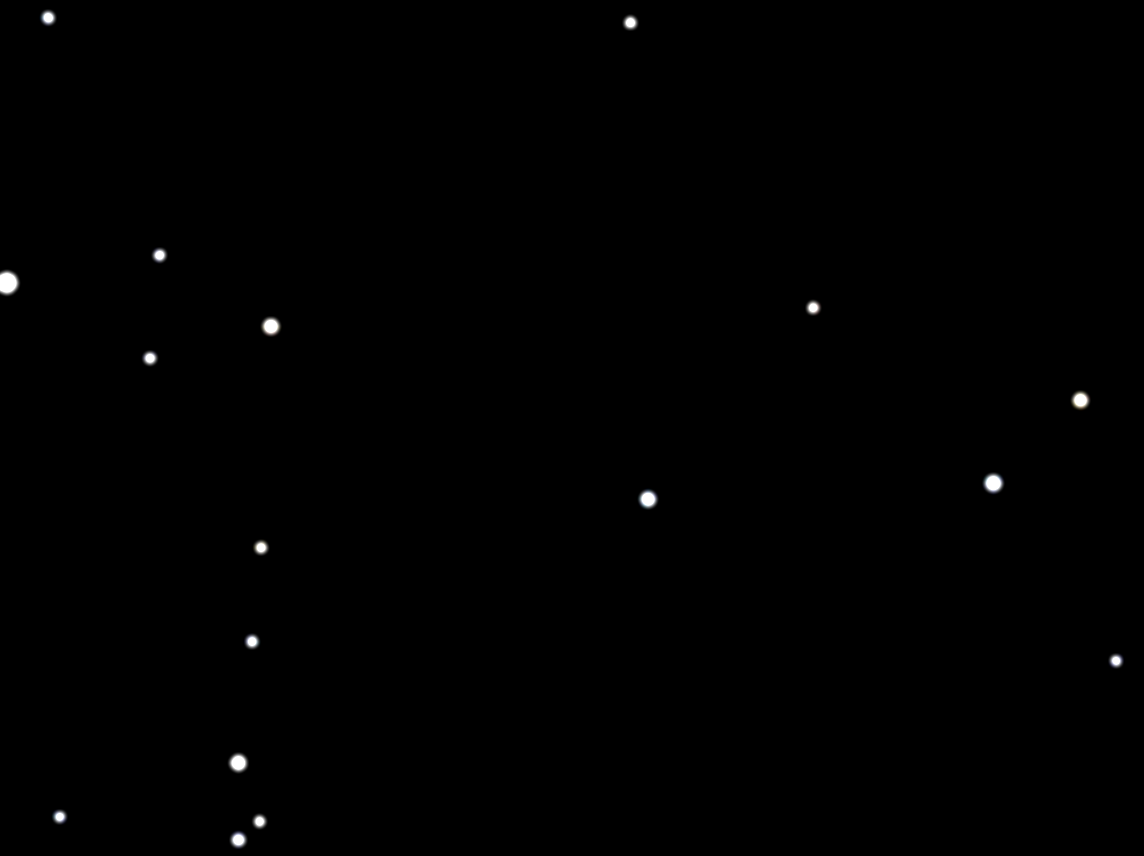}\label{fig:fov}}
\subfigure[]{\includegraphics[width=0.475\textwidth]{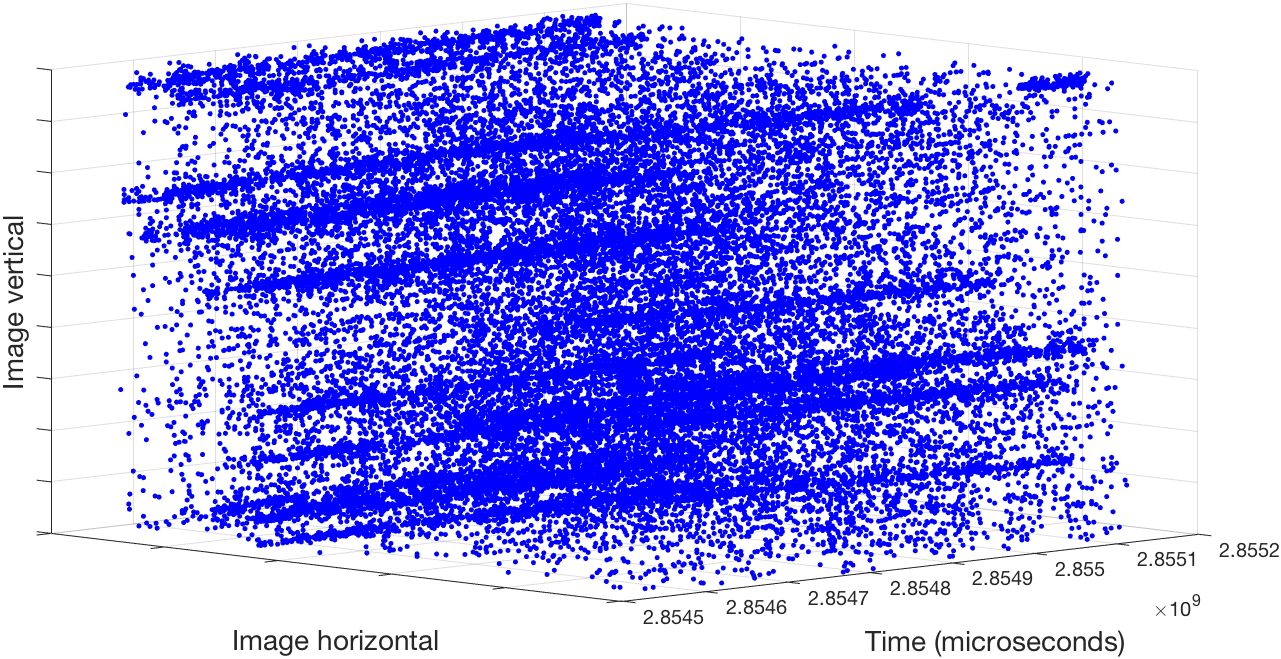}\label{fig:eventdata}}
\subfigure[]{\includegraphics[width=0.265\textwidth]{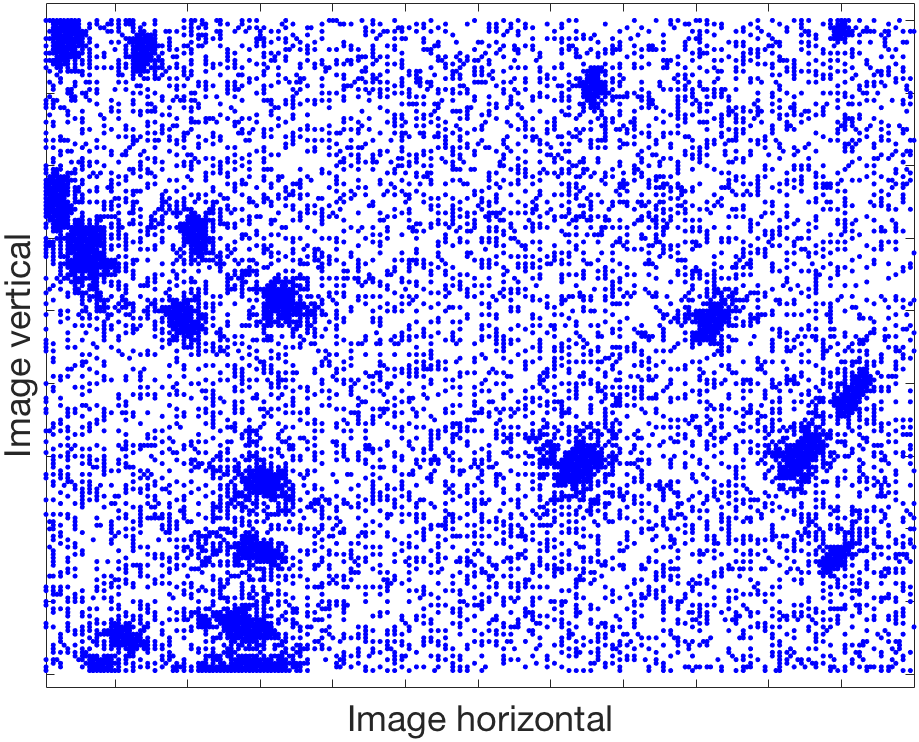}\label{fig:eventimg0}}
\caption{(a) A star field under the field of view (FOV) of a telescope. (b) Events recorded using an iniVation Davis 240C event camera as the FOV moves away from panel a (for clarity, the event polarities are not displayed here). (c) The event data collapsed onto the image frame - the star patterns in panel a are clearly observed here (though the stars are blurred due to the motion of the camera).}
\end{figure*}

Given the correspondences, the camera orientation (defined by rotation $\bR$) is computed via Wahba's problem~\cite{wiki:wahba}
\begin{align}\label{eq:wahba}
\argmin_{\bR \in SO(3)}~\sum_{p=1}^N \| \vv{\bx}_p - \bR\bX_p \|^2_2,
\end{align}
where $\vv{\bx}_p$ is the backprojected unit ray of $\bx_p$, i.e.,
\begin{align}
\vv{\bx}_p = \frac{\bK^{-1}\bar{\bx}_p}{\|\bK^{-1}\bar{\bx}_p \|_2},
\end{align}
$\bar{\bx}_p = [\bx^T_p~1]^T$, and $\bK \in \mathbb{R}^{3\times 3}$ is the camera intrinsic matrix. Many algorithms exist for solving~\eqref{eq:wahba}, such as the SVD method~\cite[Chap.~5]{markley14}. Robust versions of~\eqref{eq:wahba} also exist to deal with false correspondences~\cite{chin14}. Also critical to star tracking is a filtering step~\cite[Chap.~6]{markley14} that takes a sequence of attitude estimates to produce more refined results.

%-------------------------------------------------------------------------

\subsection{Our contributions}\label{sec:contributions}

We investigate the usage of \emph{event cameras} for attitude estimation based on star tracking. Unlike a conventional optical sensor (e.g., CCD), an event sensor detects intensity changes asynchronously~\cite{lichtsteiner08}. The output of an event camera is a set of events $\{ (\bx,t,p) \}$, where $\bx$ are the 2D coordinates of an event on the image plane, $t$ is the time of the event, and $p \in \{+,- \}$ is the polarity of the event. Fig.~\ref{fig:eventdata} illustrates event data from observing a star field.

There are two potential benefits of using event cameras for star tracking. First, due to the pausity of the scene (relatively few bright spots against a black background), the number of events generated tends to be small relative to the number of pixel positions. Hence, an event camera may consume less power. Second, event sensors have high temporal resolution (e.g., iniVation Davis 240C has $\mu s$ resolution), which could enable higher-speed star tracking. This may be useful for ultra-fine attitude control.

In this work, we do not focus on demonstrating the above benefits, since the gains from using the sensor must be put in the context of other systems or processing requirements - a complex issue that is beyond the scope of this paper. Instead, our focus is on developing an algorithmic pipeline for star tracking using event cameras, so as to establish the feasibility and promise of the paradigm.

Due to the different sensing principle, existing methods~\cite{spratling09,markley14} cannot be directly applied. Moreover, event data can be significantly noisier than conventional cameras; see Fig.~\ref{fig:eventdata}. To deal with this issue, we develop a novel processing pipeline that includes new formulations of \emph{rotation averaging} and \emph{bundle adjustment} for star tracking. We also release our event data to spark further research.

\subsection{Previous work}

Event cameras and more generally event-based processing are receiving significant attention in robotics and computer vision~\cite{github:event}. Many core capabilities, such as optic flow computation, 3D reconstruction, SLAM and visual servoing (details in~\cite{github:event}), have proven to be feasible with event cameras. In particular, in problems where high-speed operation is essential, techniques using event cameras outperform equivalent methods that use conventional cameras~\cite{mueggler14}.

Recently, there have been a few works that applied event cameras in space. In~\cite{cohen17}, the feasibility of using event cameras (aided by a telescope) to observe objects in space was established. This was followed by~\cite{cheung18}, where a probabilistic multiple hypothesis tracker (PMHT) was used to track the objects through time. These works have not described event-based star tracking for high-speed attitude estimation.

\begin{figure*}[t]
\subfigure{\includegraphics[width=0.24\textwidth]{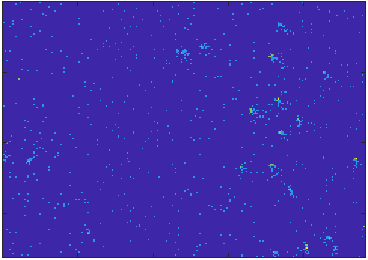}}
\hfill
\subfigure{\includegraphics[width=0.24\textwidth]{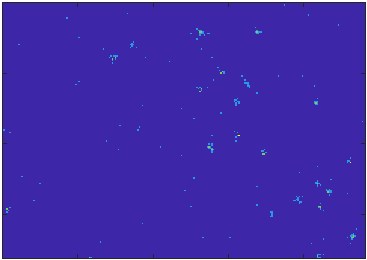}}
\hfill
\subfigure{\includegraphics[width=0.24\textwidth]{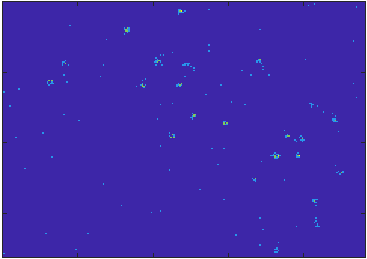}}
\hfill
\subfigure{\includegraphics[width=0.24\textwidth]{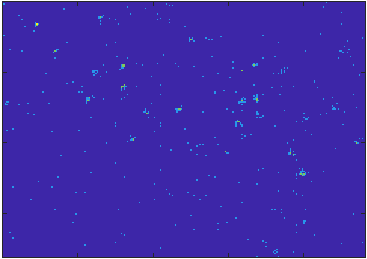}}
\subfigure{\includegraphics[width=0.24\textwidth]{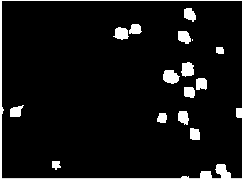}}
\hfill
\subfigure{\includegraphics[width=0.24\textwidth]{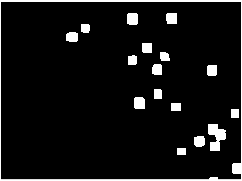}}
\hfill
\subfigure{\includegraphics[width=0.24\textwidth]{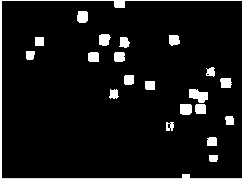}}
\hfill
\subfigure{\includegraphics[width=0.24\textwidth]{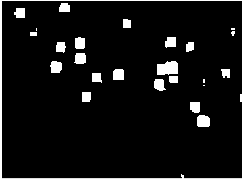}}
\caption{Top row: Event images generated from star field events according to Sec.~\ref{sec:setting}. The integration time $(t^{(i)}_{\text{start}} - t^{(i)}_{\text{end}})$ used here is $10$ ms. Brighter pixels indicate higher $I_i(\bx)$ values. Bottom row: Discrete point sets (each white pixel is a point) corresponding to the top row.}
\label{fig:eventimgs}
\end{figure*}

%-------------------------------------------------------------------------

\section{Problem setting}\label{sec:setting}

Consider event data $\mathcal{S} = \{(\bx,t,p)\}$ that was generated by observing a star field over a contiguous period of time $t \in [t_{\text{start}},t_{\text{end}}]$ under camera motion. Our overall goal is to estimate the attitude of the camera over the period of time. Many previous works on motion estimation using event cameras conduct (either implicitly or explicitly) some form of temporal aggregation on event data to elicit the geometric structures of the scene~\cite{cook11,kim14,mueggler14,mueggler15,kim16,rebecq17,gallego18}. Following these works, we generate event images
\begin{align}\label{eq:eventimgs}
I_1,  \dots, I_i , \dots, I_M
\end{align}
from $\mathcal{S}$, where each image
\begin{align}
I_i(\bx) &= \sum_{t \in [t^{(i)}_{\text{start}}, t^{(i)}_{\text{end}} ]} \delta(\bx,t),\\
 \delta(\bx,t) &= \begin{cases} 1 & \text{if $\exists(\bx,t,p) \in \mathcal{S}$}  \\ 0 & \text{otherwise}  \end{cases}
\end{align}
is obtained by collapsing all the events in a time window
\begin{align}
[t^{(i)}_{\text{start}}, t^{(i)}_{\text{end}} ] \subset [t_{\text{start}}, t_{\text{end}} ]
\end{align}
onto the image domain. In our work, the time blocks do not overlap, i.e.,
\begin{align}
[t^{(i)}_{\text{start}}, t^{(i)}_{\text{end}} ] \cap [t^{(j)}_{\text{start}}, t^{(j)}_{\text{end}} ] = \emptyset \;\;\;\; \forall i\ne j,
\end{align}
and they uniformly partition the recording duration
\begin{align}
[t^{(1)}_{\text{start}}, t^{(1)}_{\text{end}} ] \cup \dots \cup [t^{(M)}_{\text{start}}, t^{(M)}_{\text{end}} ] = [t_{\text{start}}, t_{\text{end}} ].
\end{align} 
Fig.~\ref{fig:eventimgs} (top row) illustrates.

Due to the nature of our scene (i.e., star fields), the simple temporal aggregation method above is sufficient to elicit the scene structure (cf.~the technique in~\cite{mueggler15b} that is targeted at more complex scenes). However, by comparing Figs.~\ref{fig:fov} and Fig.~\ref{fig:eventimgs} (top row), it is evident that event images are considerably noisier than conventional images. In Sec.~\ref{sec:attest}, we will describe the proposed algorithm that takes as input a sequence of noisy event images $I_1,\dots,I_i,\dots,I_M$ to compute accurate camera orientations $\hat{\bR}_1,\dots,\hat{\bR}_i,\dots,\hat{\bR}_M$.

\subsection{Why use event images?}

By generating event images, we have effectively converted $\mathcal{S}$ into a set of image frames. We emphasise that this does not defeat the purpose of using an event camera \emph{in our target problem}, since the time windows $[t^{(i)}_{\text{start}}, t^{(i)}_{\text{end}} ]$ are small (only $40$ms each), which still enables high-speed ($25$Hz) attitude estimation. In effect, we are using the event camera primarily as a high-speed low-power optical sensor.

Ideally, event data should be processed using asynchronous or event-based algorithms (e.g., the contrast maximisation framework~\cite{gallego18}) to realise the full benefit of asynchronous sensors. We leave this as future work.

\section{Attitude estimation from event images}\label{sec:attest}

\begin{figure*}[ht]\centering
\includegraphics[width=0.99\textwidth]{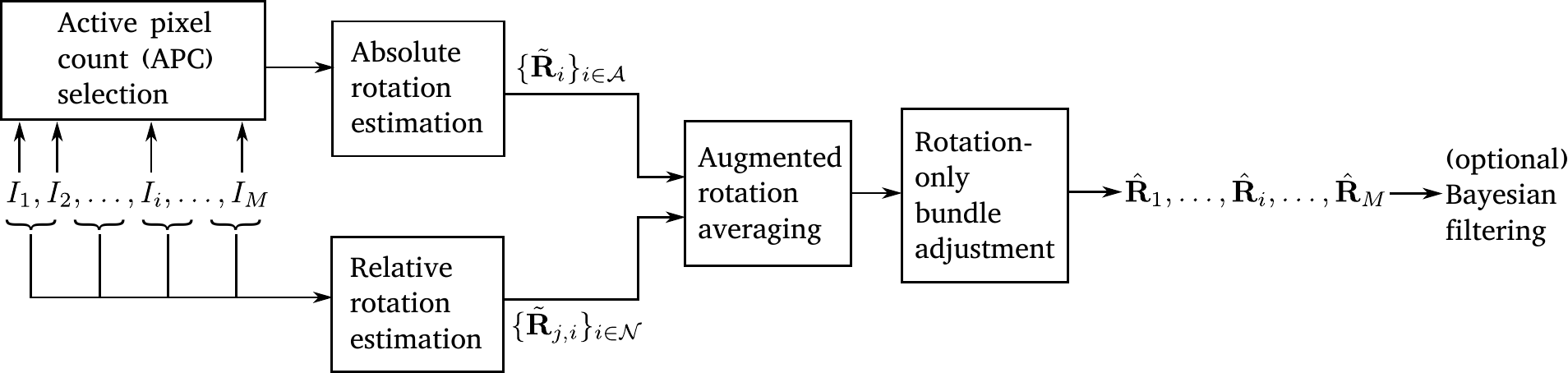}
\caption{Pipeline of proposed star tracking algorithm: given a sequence of noisy event images $I_1,\dots,I_i,\dots,I_M$, a sequence of attitude estimates $\hat{\bR}_1,\dots,\hat{\bR}_i,\dots,\hat{\bR}_M$ are produced. The output sequence can optionally be postprocessed using Bayesian filtering~\cite[Chap.~6]{markley14}.}
\label{fig:pipeline}
\end{figure*}

Fig.~\ref{fig:pipeline} shows the proposed processing pipeline. Details will be described in the rest of this section.

\subsection{Camera calibration}\label{sec:ray}

Notwithstanding the fundamentally different sensing technology, the pinhole imaging model applies to event cameras~\cite{delbruck2010}. Hence a pixel position $\bx$ in an event image $I_i$ can be backprojected to form a 3D ray by $\bK^{-1}\bar{\bx}$. For better flow, we will discuss the estimation of $\bK$ in Sec.~\ref{sec:calibration}. For now we assume $\bK$ is known without loss of generality.

\subsection{Rotation measurements}\label{sec:rotmeas}

The proposed pipeline generates and uses two types of rotation measurements from the event images: absolute rotations and relative rotations.

\subsubsection{Absolute rotation estimation}

Ideally, the processing could be simplified if we are able to estimate the attitude of each $I_i$, by performing star identification~\cite{spratling09} followed by solving Wahba's problem~\eqref{eq:wahba}. However, this is infeasible for two reasons:
\begin{itemize}[leftmargin=1em,parsep=1pt,topsep=1pt,itemsep=1pt]
\item Since most of the event images are noisy, the accuracy of star identification and attitude estimation will be poor.
\item Star identification is a relatively costly process. For example, the desktop version of~\cite{lang10,astrometry} requires seconds to process an image of size $240\times180$. It will thus be infeasible to execute star identification at event frame-rate, especially on a resource-constrained space platform.
\end{itemize}
To solve the above two difficulties, a simple heuristic (which we call \emph{active pixel count} or APC) is conducted to select only high-quality event images for star identification.

On each event image $I_i$, mean filtering by convolution with a $3\times3$ averaging kernel is first conducted. Then the APC of $I_i$ (using the mean filtered version) is calculated as
\begin{align}
APC(I_i) = \sum_{\bx} \mathbb{I}\left(I_i(\bx) \ge \epsilon_1 \right),
\end{align}
where $\epsilon_1$ is a constant threshold. Basically, $APC(I_i)$ gives the number of pixels that are ``active" (i.e., have produced a sufficient number of events) in the time slab $[t^{(i)}_{\text{start}}, t^{(i)}_{\text{end}} ]$. If
\begin{align}
APC(I_i) \ge \epsilon_2,
\end{align}
where $\epsilon_2$ is another constant threshold, then we regard $I_i$ as suitable for direct attitude estimation. The heuristic is based on the observation that brighter and more well-defined stars tend to yield event data with more active pixels.

Let $\cA \subset \{1,\dots,M \}$ index the event images that are selected via the APC heuristic. To ensure accurate attitude estimates from this set, we use relatively high values for $\epsilon_1$ and $\epsilon_2$ (in our experiments, $\epsilon_1 = 2$ and $\epsilon_2 = 50$), thus, $\cA$ tends to be small, i.e., $|\cA| \ll M$. We subject set $\cA$ to star identification (we applied~\cite{lang10} as implemented in~\cite{astrometry} in our pipeline), followed by the SVD technique for Wahba's problem~\cite{wiki:wahba}, to yield a set of attitude estimates
\begin{align}\label{eq:abrot}
\{ \tilde{\bR}_i \}_{i \in \cA},
\end{align}
We call these rotation measurements the \emph{absolute rotations}.

%As shown in Fig.~\ref{fig:eventimgs} (top row), the event images are able to capture the geometric patterns of the stars in the FOV. In fact, the event images can be directly subjected to a star identification algorithm for attitude determination. However, star identification is a relatively costly process; for example, based on our testing, the desktop version of~\cite{astrometry} requires time in the order of seconds to process an image\footnote{This can be speeded up by ``warm starting" the star map search with an approximate orientation. However, in the context of a state estimation problem, this has the danger of biasing the outcomes.}, which makes it impractical to execute it on every event image, especially on a constrained space platform.

\subsubsection{Relative rotation estimation}

To fully make use of available data, we estimate \emph{relative rotations} from the event images. Let $I_i$ and $I_j$ be two event images that overlap, i.e., they observe some common stars in their respective FOVs. The relative rotation $\bR_{j,i}$ between the images aligns the noisy coordinates $\bx$ and $\bx^\prime$ of a star that is observed simultaneously in $I_i$ and $I_j$, i.e.,
\begin{align}
\vv{\bx}^\prime \approx \bR_{j,i}\vv{\bx}.
\end{align}
Note that since stars are at infinity, there is no parallax between $I_i$ and $I_j$, which justifies rotational alignment~\cite{markley14}.

To estimate $\bR_{j,i}$, we begin by thresholding on the mean filtered versions of $I_i$ and $I_j$ to create discrete point sets
\begin{align}\label{eq:pointsets}
\{ \bx_{i,p} \}^{P}_{p=1} \;\;\; \text{and} \;\;\; \{ \bx_{j,q} \}^{Q}_{q=1}.
\end{align}
See Fig.~\ref{fig:eventimgs} (bottom row) for example point sets generated. Note that at this stage, data association has not been performed and $P \ne Q$ in general. %The aim of local averaging is to suppress the spurious (random) events in the event images; cf.~bottom and top rows in Fig.~\ref{fig:eventimgs}. 

Given the point sets~\eqref{eq:pointsets}, to obtain the relative rotation measurement $\tilde{\bR}_{j,i}$ we solve the registration problem
\begin{align}\label{eq:ticp}
\tilde{\bR}_{j,i} = \argmin_{\bR \in SO(3)}~\sum^{L}_{p=1} r_{(p)}(\bR),
\end{align}
where $L < P$. The per-point error is computed as
\begin{align}
r_{p}(\bR) = \min_{q}\left\| \bR \vv{\bx}_{i,p} - \vv{\bx}_{j,q} \right\|_2,
\end{align}
and $r_{(p)}(\bR)$ is the $p$-th largest value of the set
\begin{align}
\{ r_{p}(\bR) \}_{p=1}^P
\end{align}
By setting $L < P$, we obtain a robust solution since only the $L$-smallest residuals are minimised. In other words, points from $I_i$ due to spurious events that have no correspondences in $I_j$ (i.e., the outliers) have no effect on the solution. Problem~\eqref{eq:ticp} can be solved efficiently by a variant of ICP called \emph{trimmed ICP}; see~\cite{chetverikov02} for details of the algorithm.

It is computationally wasteful to attempt~\eqref{eq:ticp} on all pairs of event images, hence, we only conduct trimmed ICP on image pairs that are within a time window of fixed size $W$ (we set $W = 5$ in our experiments). This creates a set of relative rotation measurements
\begin{align}\label{eq:relrot}
\{ \tilde{\bR}_{j,i} \}_{\langle j,i \rangle \in \mathcal{N}},
\end{align}
where $\cN$ is the adjacency graph such that $\langle j,i \rangle \in \mathcal{N}$ if and only if $|i-j| \le W$.

\subsection{Optimisation}\label{sec:optimisation}

The aim of optimisation is to denoise and fuse the different rotation measurements to yield accurate attitude estimates. This is accomplished efficiently using two novel formulations of rotation averaging and bundle adjustment.

\subsubsection{Augmented rotation averaging}

The original form of rotation averaging~\cite{hartley13} takes as input a set of relative rotations and computes a set of absolute rotations. Put in our context, this is the problem
\begin{align}\label{eq:stdrotavg}
\min_{ \{ \bR_i \}_{i = 1}^M}~\sum_{\langle j,i \rangle \in \cN} \left\| \bR_j -  \tilde{\bR}_{j,i} \bR_i  \right\|_F,
\end{align}
where $\|~\|_F$ is the Frobenius norm. However, this formulation is unsuitable since $\cN$ is a chain with no loops, and optimising the attitudes using~\eqref{eq:stdrotavg} will lead to drift errors.

To mitigate drift error, the absolute rotations~\eqref{eq:abrot} must be factored into rotation averaging. To this end, we formulate the \emph{augmented rotation averaging} problem
\begin{align}\label{eq:augrotavg}
\begin{aligned}
& \underset{\{ \bR_i \}^{M}_{i=1}, \bR_{M+1}}{\min}
& & \sum_{\langle j,i \rangle \in \cN} \left\| \bR_j -  \tilde{\bR}_{j,i} \bR_i  \right\|_F\\
& & & + \alpha \sum_{k \in \cA} \left\| \bR_k - \tilde{\bR}_k \bR_{M+1} \right\|_F \\
& \text{s.t.}
& & \bR_{M+1} = \bI,
\end{aligned}
\end{align}
where $\bR_{M+1}$ is a ``dummy" attitude variable, $\bI$ is the identity matrix, and $\alpha$ is a positive constant that defines the relative importance of the relative and absolute rotations. Intuitively, adding error terms of the form
\begin{align}
\left\| \bR_k - \tilde{\bR}_k \bR_{M+1} \right\|_F = \left\| \bR_k - \tilde{\bR}_k \right\|_F, \;\;\;\; k \in \cA
\end{align}
encourage consistency between some of the attitude estimates and the measured absolute rotations, which is then propagated to the rest of the sequence. 

Despite having the same form as~\eqref{eq:stdrotavg} except for the constraint $\bR_{M+1} = \bI$, existing rotation averaging algorithms~\cite{martinec07,moulon13,hartley13,chatterjee2013efficient,arrigoni14,carlone15} (which are tailored for~\eqref{eq:stdrotavg}) cannot be directly applied to~\eqref{eq:augrotavg}. A simple workaround is as follows: temporarily ignore the constraint $\bR_{M+1} = \bI$ in~\eqref{eq:augrotavg} and optimise the attitudes using an existing rotation averaging algorithm (we used~\cite{chatterjee2013efficient} in our work). Then, right multiply each of the estimated attitude $\hat{\bR}_i$ with $(\hat{\bR}_{M+1})^{-1} = (\hat{\bR}_{M+1})^{T}$.

It has been shown that rotation averaging is quite insensitive to initialisations~\cite{olsson11,eriksson18}, thus, when solving~\eqref{eq:augrotavg} we simply initialise all rotation variables as the identity matrix.

\subsubsection{Rotation-only bundle adjustment}

The solutions from rotation averaging are then refined using \emph{rotation-only bundle adjustment}. First, as a by-product of computing the relative rotations~\eqref{eq:ticp}, we associate points across event images to form \emph{star tracks} $\{  \cY_s  \}_{s=1}^S$, where
\begin{align}
\cY_s = \{ \by_{i,s} \in \mathbb{R}^2 \mid \eta(i,s) = 1 \},\\
\eta(i,s) = \begin{cases} 1 & \text{if the $s$-th star is seen in $I_i$,} \\ 0 & \text{otherwise,} \end{cases}
\end{align}
and $\by_{i,s}$ are the 2D coordinates of the $s$-th observed star as viewed in event image $I_i$. We then define the nonlinear least squares (NLS) problem
\begin{align}\label{eq:ba}
\begin{aligned}
& \underset{\{ \bR_i \}, \{ \bX_s \}}{\text{min}}
& & \sum_{i,s} \eta(i,s) \left\| \vv{\by}_{i,s} - \bR_i\bX_s \right\|^2_2 \\
& \text{s.t.}
& & \bX_s \in \mathbb{R}^3~\text{and}~\| \bX_s \|_2 = 1,~\forall s,
\end{aligned}
\end{align}
where $\bX_s$ defines the 3D direction of the $s$-th observed star as seen from the inertial frame. Intuitively,~\eqref{eq:ba} is a special case of bundle adjustment~\cite{triggs1999bundle} where only rotational motion and 3D directions are optimised.

Compared to~\eqref{eq:augrotavg}, problem~\eqref{eq:ba} takes into account the observed star coordinates in the estimation. To use~\eqref{eq:ba} to refine the solutions of~\eqref{eq:augrotavg}, we simply initialise the rotation variables in~\eqref{eq:ba} using the output of~\eqref{eq:augrotavg}, then apply a standard NLS solver (i.e., Ceres~\cite{ceres-solver}) to carry out bundle adjustment. As with most bundle adjustment solvers, Ceres can take into account problem sparsity (i.e., not all points are seen in every frame) to speed up convergence.

To initialise the star directions $\{ \bX_s \}$, we compute
\begin{align}
\bX_s = \frac{1}{|\cY_s|}\sum_i \eta(i,s) \hat{\bR}^{-1}_i\vv{\by}_{i,s},
\end{align}
i.e., the mean direction of the separate observations of the $s$-th star as seen in the inertial frame, using initial estimates of the attitudes $\{ \hat{\bR}_i \}$ from rotation averaging~\eqref{eq:augrotavg}.

Secs.~\ref{sec:simul} and~\ref{sec:results} will describe the testing methodology and results of the proposed star tracking approach.

\section{Generating testing data}\label{sec:simul}

Conducting a space mission for the purpose of our paper is beyond our budget. Following other works in astronautics research (e.g.,~\cite{opromolla2017new}), we use simulation data to test our method. While it is possible to use the event data simulator by Mueggler et al.~\cite{mueggler17}, for our problem it generates unrealistically clean data due to the simplicity of the scene (points at infinity with no other structures or occlusions).

\begin{figure}[t]\centering
\subfigure[]{\includegraphics[width=0.6\columnwidth]{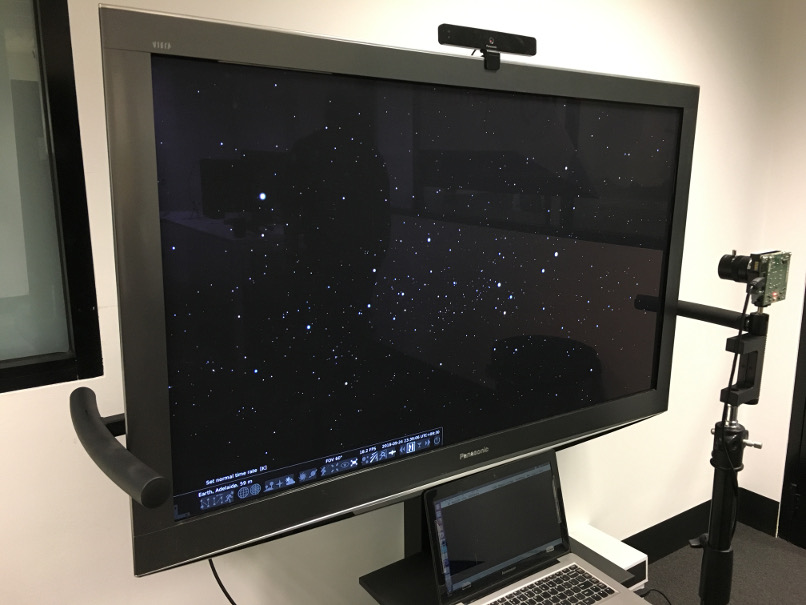}\label{fig:setup}}\\
\vspace{-1em}
\subfigure[]{\includegraphics[width=1.0\columnwidth]{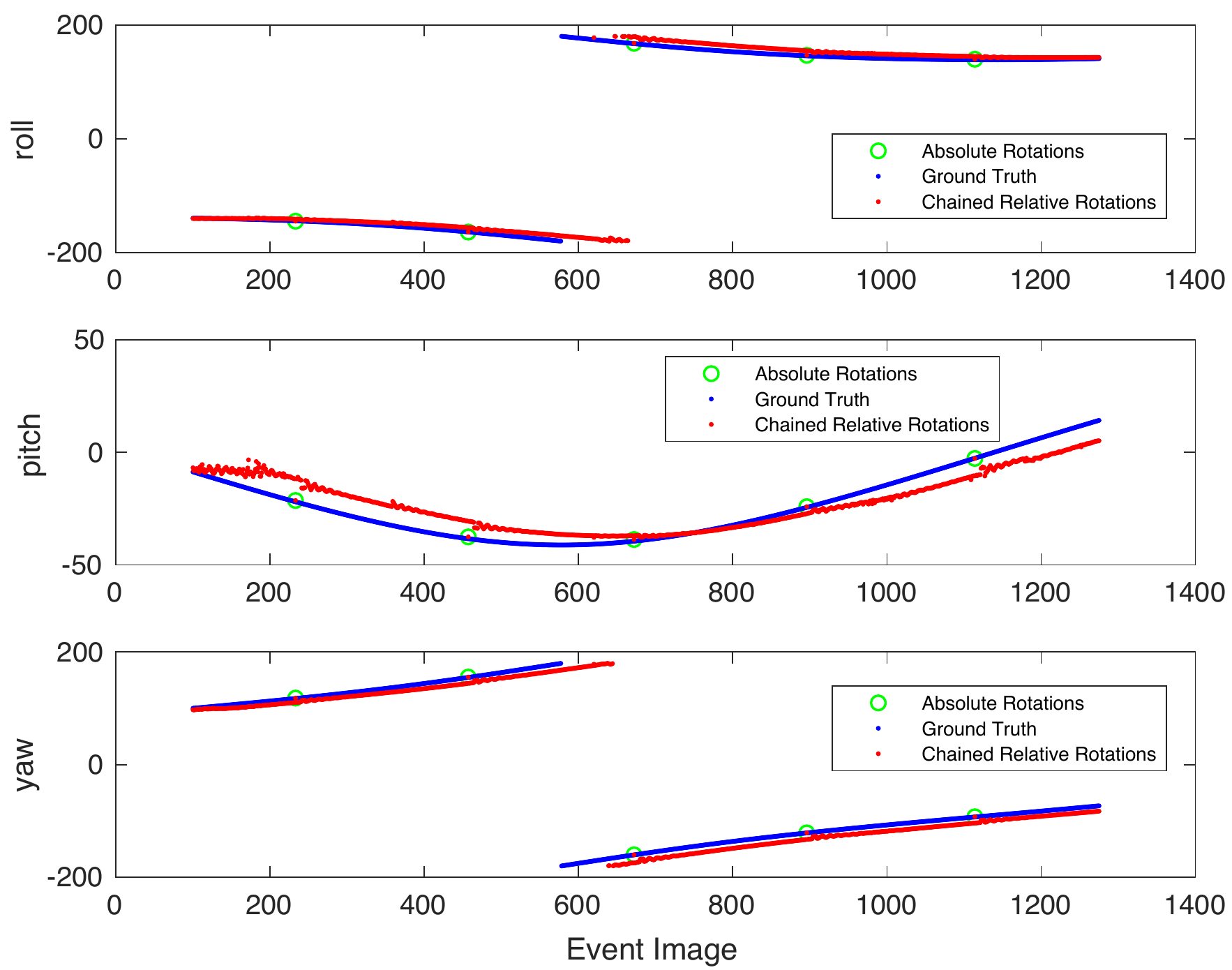}\label{fig:datagen}}
\caption{(a) Testing data generation rig. (b) Ground truth attitudes and initial attitude estimates (both sets expressed as Euler angles~\cite{wiki:euler}) for a sample event data sequence generated according to Sec.~\ref{sec:simul}. See Sec.~\ref{sec:errorAnalysis} on deriving the initial attitudes from the rotation measurements computed according to Sec.~\ref{sec:rotmeas}.}
\end{figure}

The approach we have taken is to use the planetarium software Stellarium\footnote{\url{https://stellarium.org}} to render real star fields on a screen, then capture the screen using an event camera, specifically the iniVation Davis 240C. Fig.~\ref{fig:setup} shows our setup. Since our target operating environment is space, where the atmosphere is much thinner, stars imaged in space will have constant brightness. This supports the usage of Stellarium which does not simulate atmospheric effects.

Given a fixed FOV and an arbitrary initial attitude, a rotational motion of a fixed angular velocity is executed in Stellarium to continuously change the attitude and generate event data. Since the motion is controlled, the ground truth attitudes throughout the recording duration can be analytically calculated; Fig.~\ref{fig:datagen} illustrates ground truth attitudes for such a sequence. In the above way, we can generate event data sequences for objective testing. Before further describing the experiments and results in Sec.~\ref{sec:results}, we first explain other preprocessing and calibration steps.

\subsection{Removing spurious events}

In our setup, the events are mainly triggered by two sources of intensity changes: the star motions and the screen refresh. Events due to the latter  are considered spurious in our setting, thus we removed them using the Spatial Bandpass Filter tool in the jAER suite\footnote{\url{https://github.com/SensorsINI/jaer}}. Of course there remain other noisy events due to inherent flaws of the sensor.

\subsection{Calibrating the virtual telescope}\label{sec:calibration}

Mathematically, Stellarium acts as a ``virtual telescope". By imaging the screen output of Stellarium with an event camera, we can relate a pixel position $\bx$ on the event camera and the corresponding star direction $\bX$ with the equation
\begin{align}
\bar{\bx} = \bK_{\text{ev}}\mathbf{H}_{\text{sc}}\bK_{\text{te}}\bR\bX \equiv \bK\bR\bX,
\end{align}
where $\bK_{\text{ev}}$, $\mathbf{H}_{\text{sc}}$, $\bK_{\text{te}}$ and $\bR$ (all $3 \times 3$) are defined as follows:
\begin{itemize}[leftmargin=1em,itemsep=1pt,topsep=1pt,parsep=1pt]
\item $\bK_{\text{ev}}$ is the intrinsic matrix of the event camera.
\item $\mathbf{H}_{\text{sc}}$ is the homography that accounts for the non-fronto-parallel viewpoint of the event camera to the screen. A homography is mathematically valid since there is no parallax in star field images, and the screen and event camera image plane differ by a planar perspective transform.
\item $\bK_{\text{te}}$ is the intrinsic matrix of the virtual telescope (determines the magnification of the instrument etc.).
\item $\bR$ is the attitude of the telescope (the quantity of interest).
\end{itemize}
Calibrating for the overall intrinsic matrix $\bK$ (cf.~Sec.~\ref{sec:ray}) is thus achieved by calculating $\bK_{\text{ev}}$, $\mathbf{H}_{\text{sc}}$ and $\bK_{\text{te}}$.

The intrinsic matrix $\bK_{\text{ev}}$ can be estimated using existing techniques; see~\cite[Calibration]{github:event}. The homography $\mathbf{H}_{\text{sc}}$ can also be estimated using standard methods~\cite{hartley04} given sufficient 2D-2D correspondences (manually extracted) between an event image and the screen output of Stellarium.

To compute $\bK_{\text{te}}$, define the pinhole projection
\begin{align}
\bar{\bz} = \bK_{\text{te}}\bR\bX \equiv \bP\bX
\end{align} 
performed by Stellarium at a particular magnification, where $\bz$ are the image coordinates of a star direction $\bX$. Given 2D-3D correspondences $\{(\bz_i,\bX_i)\}^{N}_{i=1}$ between a set of observed stars in the Stellarium image and their corresponding star directions (the latter can be easily retrieved via the software interface), we set up the linear system
\begin{align}
\left[\begin{matrix} \bar{\bz}_1 \\ \vdots \\ \bar{\bz}_N \end{matrix} \right] = \bP \left[ \begin{matrix} \bX_1 \\ \vdots \\ \bX_N \end{matrix} \right]
\end{align}
and solve for $\bP$ using standard linear least squares. Then, $\bK_{\text{te}}$ and $\bR$ can be obtained from $\bP$ via QR decomposition.

\section{Results}\label{sec:results}

Using the methodology for generating testing data in Sec.~\ref{sec:simul}, we generated a number of event data sequences to test our star tracking pipeline.  In all sequences, we fixed the FOV on Stellarium to 20 degrees, the recording duration to 45 seconds, and the angular velocity to $4^\circ/s$. The integration time $t^{(i)}_{\text{end}}  - t^{(i)}_{\text{start}}$ for event image generation was fixed at $40ms$ for all sequences, hence there are $1125$ event images per sequence.  In all the testing sequences, the initial attitude and rotational motion were chosen arbitrarily.

In this section, we show the results on six of the generated sequences; see the supplementary material for videos of the corresponding event images (results on more sequences are also available there), and~\cite{site} to obtain the data.

\subsection{Error analysis of input rotations}\label{sec:errorAnalysis}

On each testing sequence, via Sec.~\ref{sec:rotmeas}, we obtained absolute rotations $\{ \tilde{\bR}_i \}_{i \in \cA}$ and relative rotations $\{ \tilde{\bR}_{j,i} \}_{\langle j,i\rangle \in \cN}$. Let $\{ \bR^*_{i} \}^M_{i=1}$ be the ground truth attitudes. The angular error between two rotations $\bR_1$ and $\bR_2$~\cite{hartley13} is
\begin{align}
\angle(\bR_1,\bR_2) = 2\arcsin\left((2\sqrt{2})^{-1} \| \bR_1 - \bR_2 \|_F\right).
\end{align}
To analyse the quality of the rotation measurements:
\begin{itemize}[leftmargin=1em,itemsep=1pt,topsep=1pt,parsep=1pt]
\item Table~\ref{tab:astrometryErrTab} summarises the angular error between absolute rotations $\{ \tilde{\bR}_i \}_{i \in \cA}$ and corresponding ground truth rotations $\{ \bR^\ast_i \}_{i \in \cA}$ over the six sequences, where there are cumulatively $33$ absolute rotations. It can be seen that most of the absolute rotations are reasonably accurate, indicating the effectiveness of the APC heuristic in selecting event images for star identification.
\item The ground truth relative rotation between $I_i$ and $I_j$ is
\begin{align}
\bR^\ast_{j,i} = \bR_i^\ast (\bR_j^\ast)^T.
\end{align}
Table~\ref{tab:relativeErrTab_} displays the RMSE and standard deviation (SD) of the relative angular error $\angle(\tilde{\bR}_{j,i},\bR^\ast_{j,i})$ in each sequence. It is evident that the relative rotations are very accurate ($\le 1^\circ$ RMSE), which supports our idea of using absolute rotations sparingly (since they are costly to compute), and relative rotations extensively (see Sec.~\ref{sec:rotmeas}).
\end{itemize}

To visualise the input rotations, in Fig.~\ref{fig:datagen} we plot the attitudes that were obtained by chaining the absolute and relative rotations of the particular sequence. Note that there are multiple ways to chain the rotations, and an arbitrary chaining order was used in the figure - since chaining was done mainly for illustration (NB: our processing pipeline does not require it as initialisation), the choice is sufficient. The more important insight is that, while the \emph{individual} input rotations were accurate, simply chaining them without further optimisation will lead to significant drift error.

\subsection{Qualitative results}

\begin{figure}[t]\centering
\includegraphics[width=1.0\columnwidth]{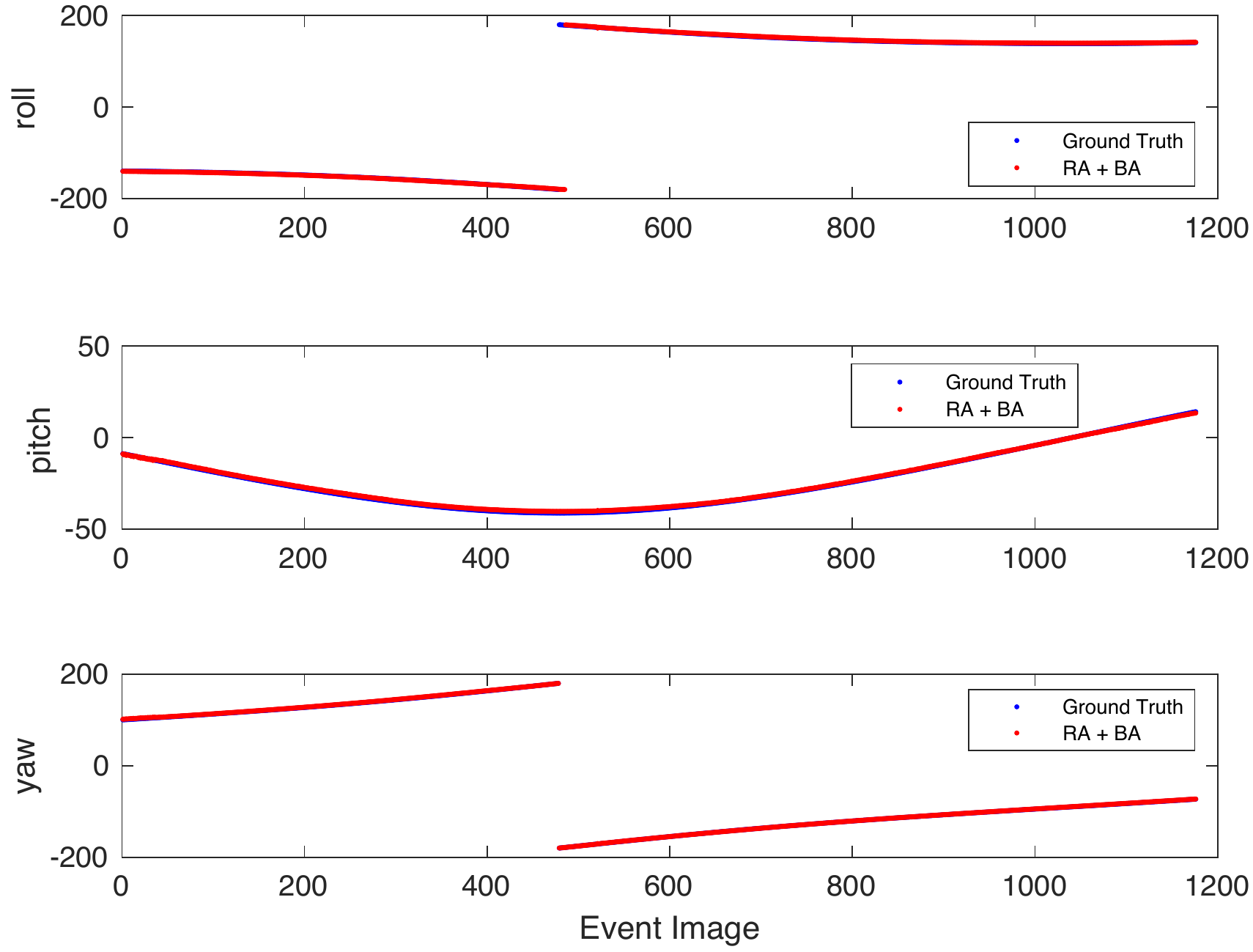}
\caption{Visualisation of optimised attitudes and ground truth attitudes for a testing sequence. Both sets of values are very close.}
\label{fig:adelaideYPR}
\end{figure}

Fig.~\ref{fig:adelaideYPR} plots the attitudes (in Euler angles) resulting from rotation averaging and bundle adjustment, and the ground truth attitudes, for the sequence in Fig.~\ref{fig:datagen}. It is evident that the proposed optimisation routines have reduced the errors significantly and accurately estimated the attitudes.

\begin{table}[t]\centering
\begin{tabular}{c c c c  }
\hline
Angular error & $< 1^\circ$ & $< 10^\circ$ & $> 10^\circ$  \\
\hline
\hline
\# absolute rotations & 30 & 1 & 2 \\
\hline
\end{tabular}
\caption{Angular error of absolute rotations $\{ \tilde{\bR}_i \}_{i \in \cA}$.}
\label{tab:astrometryErrTab}
\end{table}

\begingroup
\setlength{\tabcolsep}{3.20pt} % Default value: 6pt
\begin{table}[t]\centering
\begin{tabular}{ l c c c c c c }
\hline
Seq \# & 1 & 2 & 3 & 4 & 5 & 6 \\
\hline
\hline
RMSE & 0.0166 & 0.01276 & 0.2175 & 0.2345 & 0.3185 & 0.2234\\
SD & 0.1239 & 0.155 & 0.2525 & 0.3423 & 0.1421 & 0.3121\\
\hline
\end{tabular}
\caption{Angular error of relative rotations $\{ \tilde{\bR}_{j,i} \}_{\langle j,i\rangle \in \cN}$.}
\label{tab:relativeErrTab_}
\end{table}
\endgroup

\subsection{Quantitative results}

Fig.~\ref{fig:sixseqs} shows quantitative results for six sequences (more results in supplementary material). For each sequence, we plot the angular errors (as well as their RMSE and SD) between the ground truth attitudes and
\begin{itemize}[leftmargin=1em,itemsep=1pt,topsep=1pt,parsep=1pt]
\item The initial attitudes obtained by an arbitrary way of chaining the input rotations (again, this is done only for illustration and comparison purposes, since our algorithms do not need to be initialised in this manner);
\item The estimates after augmented rotation averaging; and
\item The estimates after augmented rotation averaging and rotation-only bundle adjustment.
\end{itemize}
In all the sequences, our pipeline achieved an error of $\le 1^\circ$ RMSE, which is on par with commercial star trackers~\cite{opromolla2017new}.

\subsection{Runtime}

Executed on a $2.9$GHz Intel i7 machine, the average total runtime per sequence is $58.8s$, which includes time for event image generation ($8.1s$), extracting rotation measurements ($41.3s$) and optimisation ($9.4s$). A simple apportionment over the $1125$ event images in each sequence gives $\approx 19$ FPS. Note that this result is mainly to indicate the efficiency of the proposed algorithms - not the speed of the star tracker. As mentioned in Sec.~\ref{sec:contributions}, the actual speed will likely depend on overall system design.

\begin{figure*}[t]\centering
\subfigure[Sequence 1]{\includegraphics[width=0.46\textwidth]{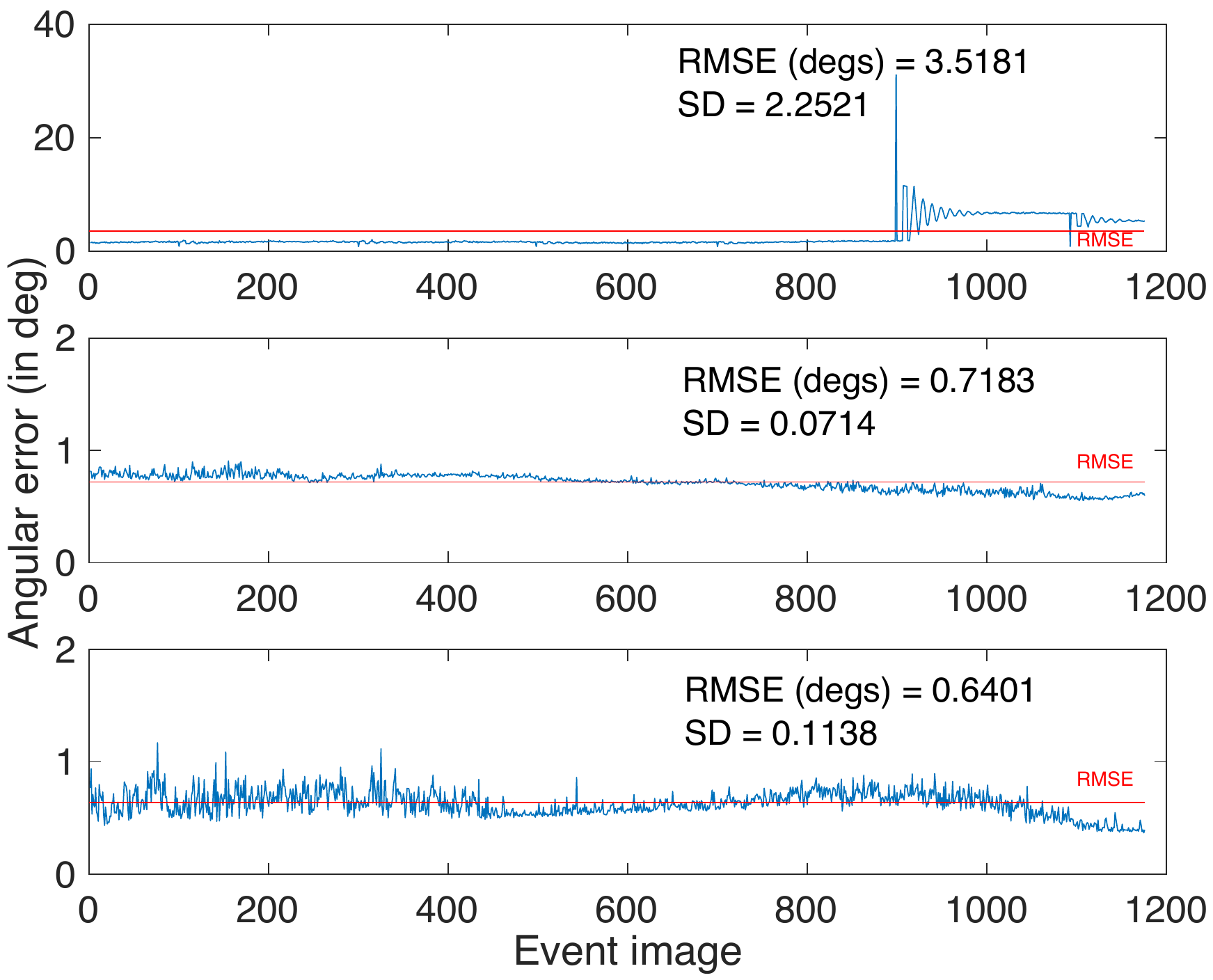}\label{fig:adelaideRABA}}\hspace{1em}
\subfigure[Sequence 2]{\includegraphics[width=0.46\textwidth]{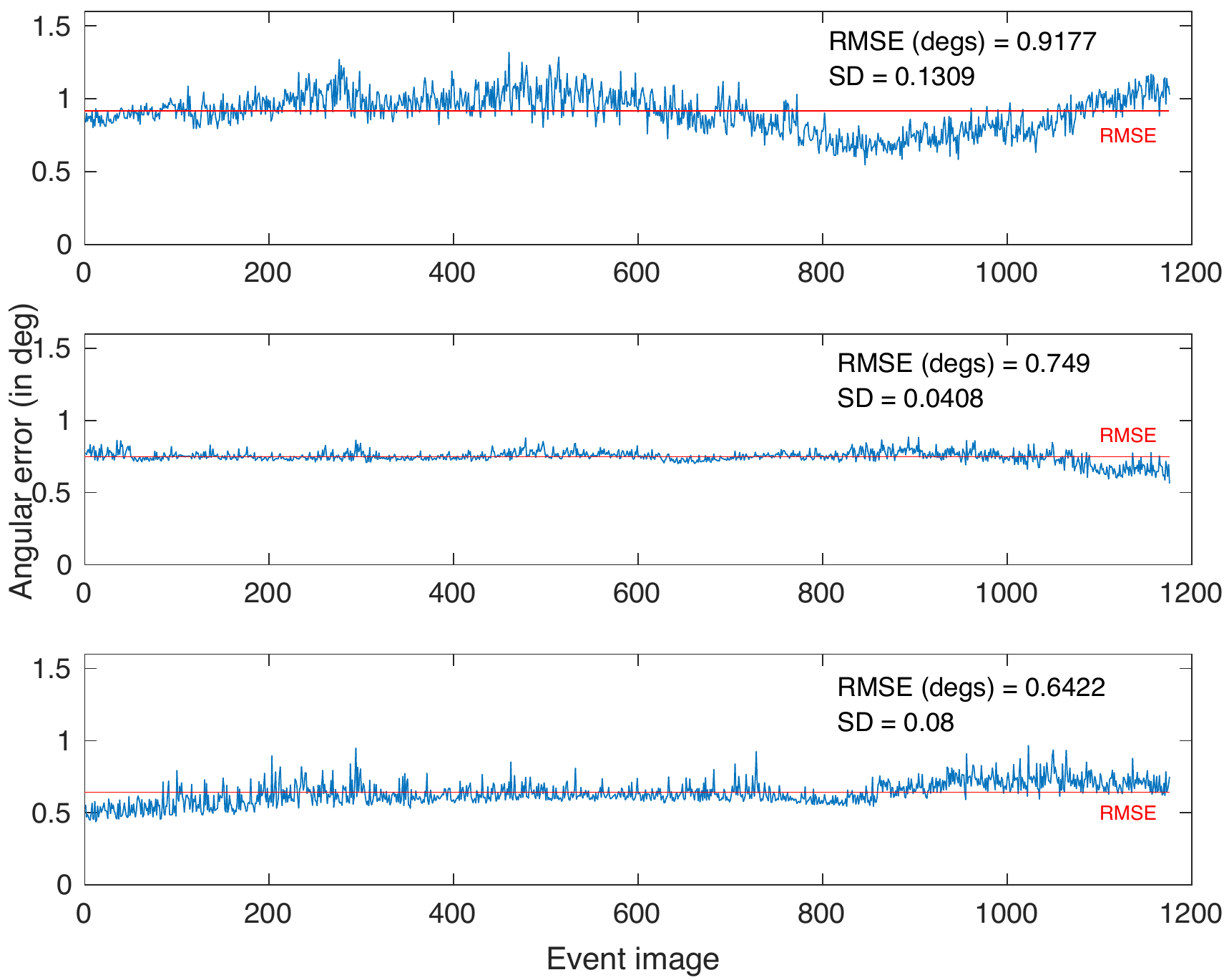}\label{fig:capetownRABA}}
\subfigure[Sequence 3]{\includegraphics[width=0.46\textwidth]{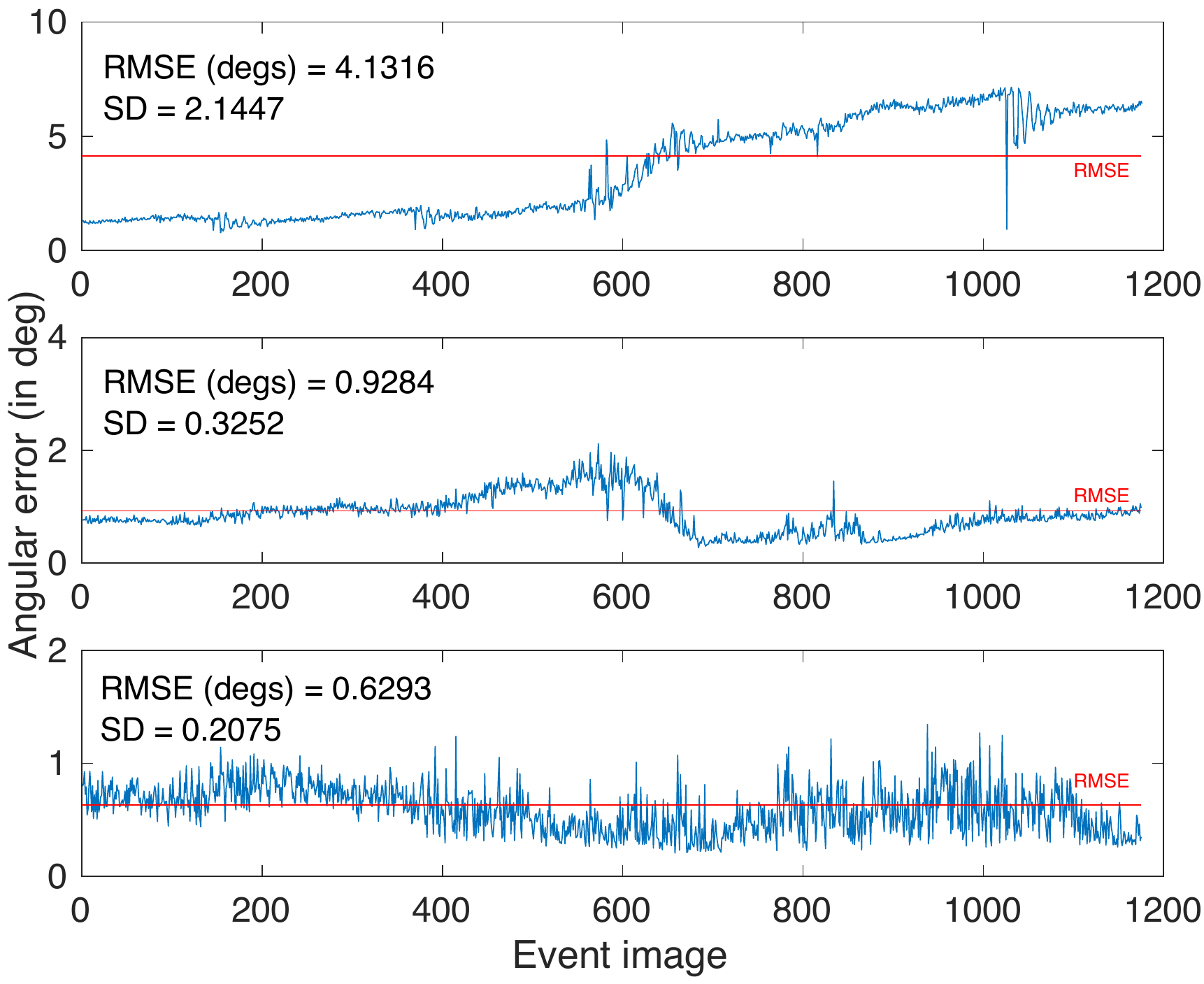}\label{fig:ecuadorRABA}}\hspace{1em}
\subfigure[Sequence 4]{\includegraphics[width=0.46\textwidth]{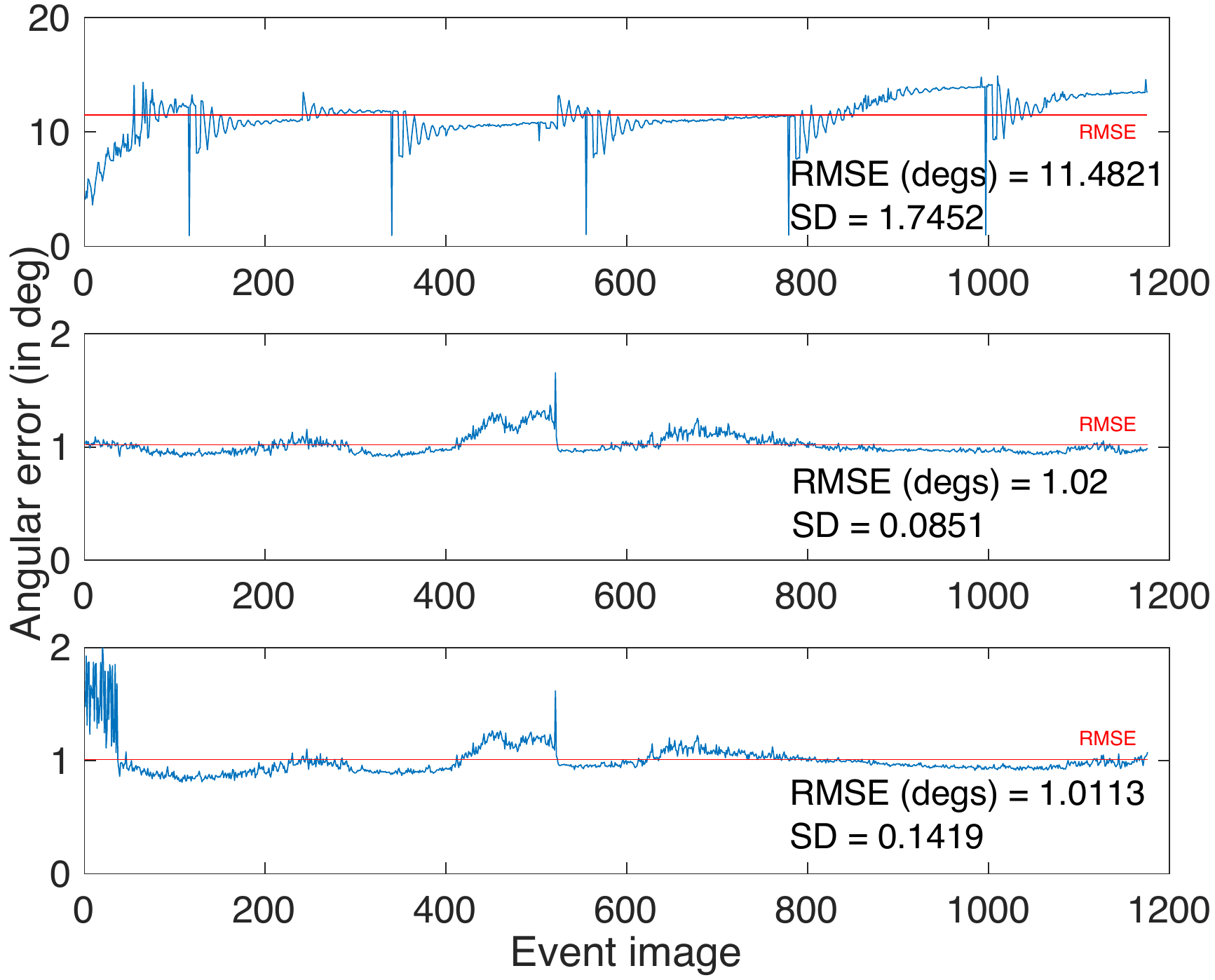}\label{fig:parisRABA}}
\subfigure[Sequence 5]{\includegraphics[width=0.46\textwidth]{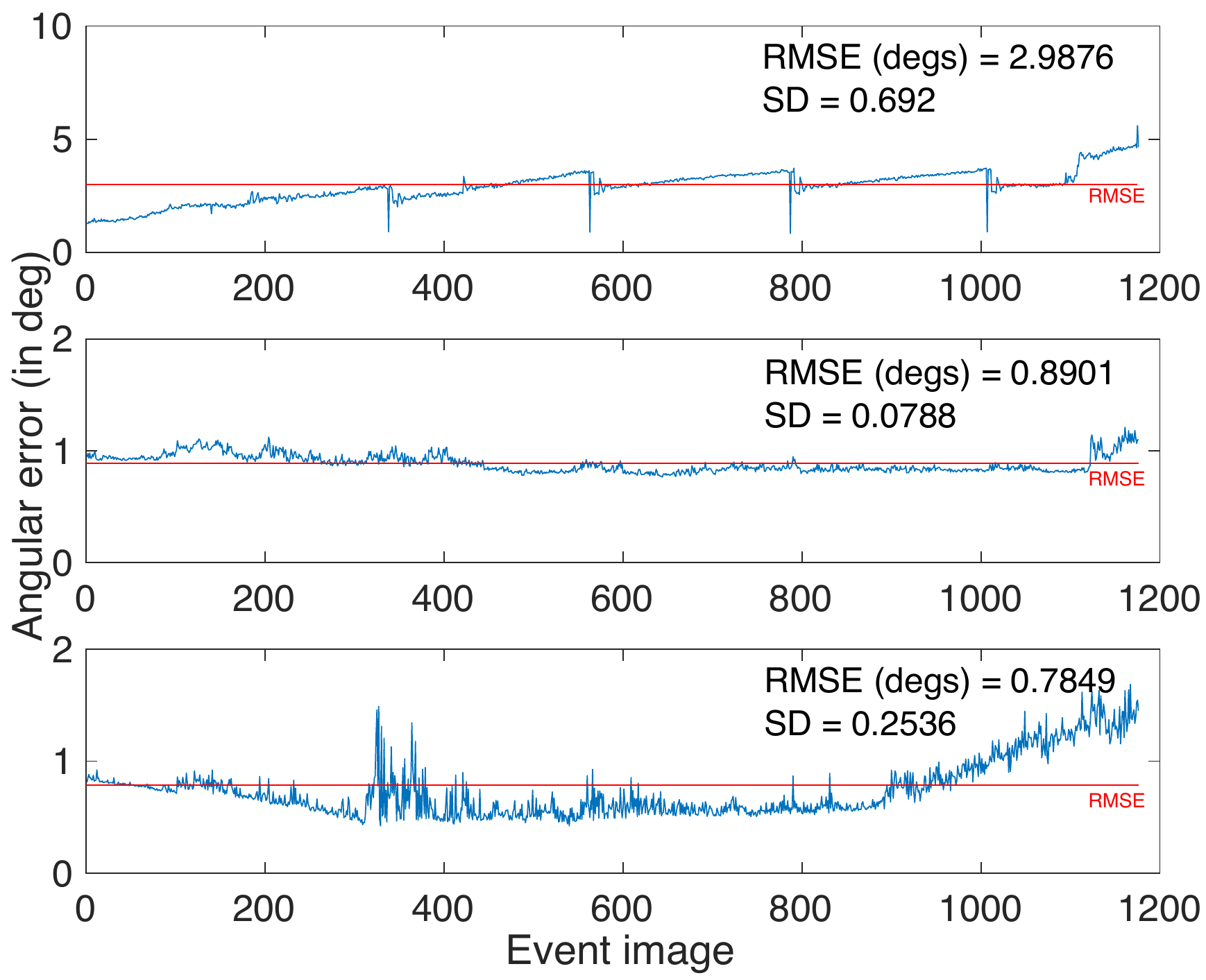}\label{fig:tokyoRABA}}\hspace{1em}
\subfigure[Sequence 6]{\includegraphics[width=0.46\textwidth]{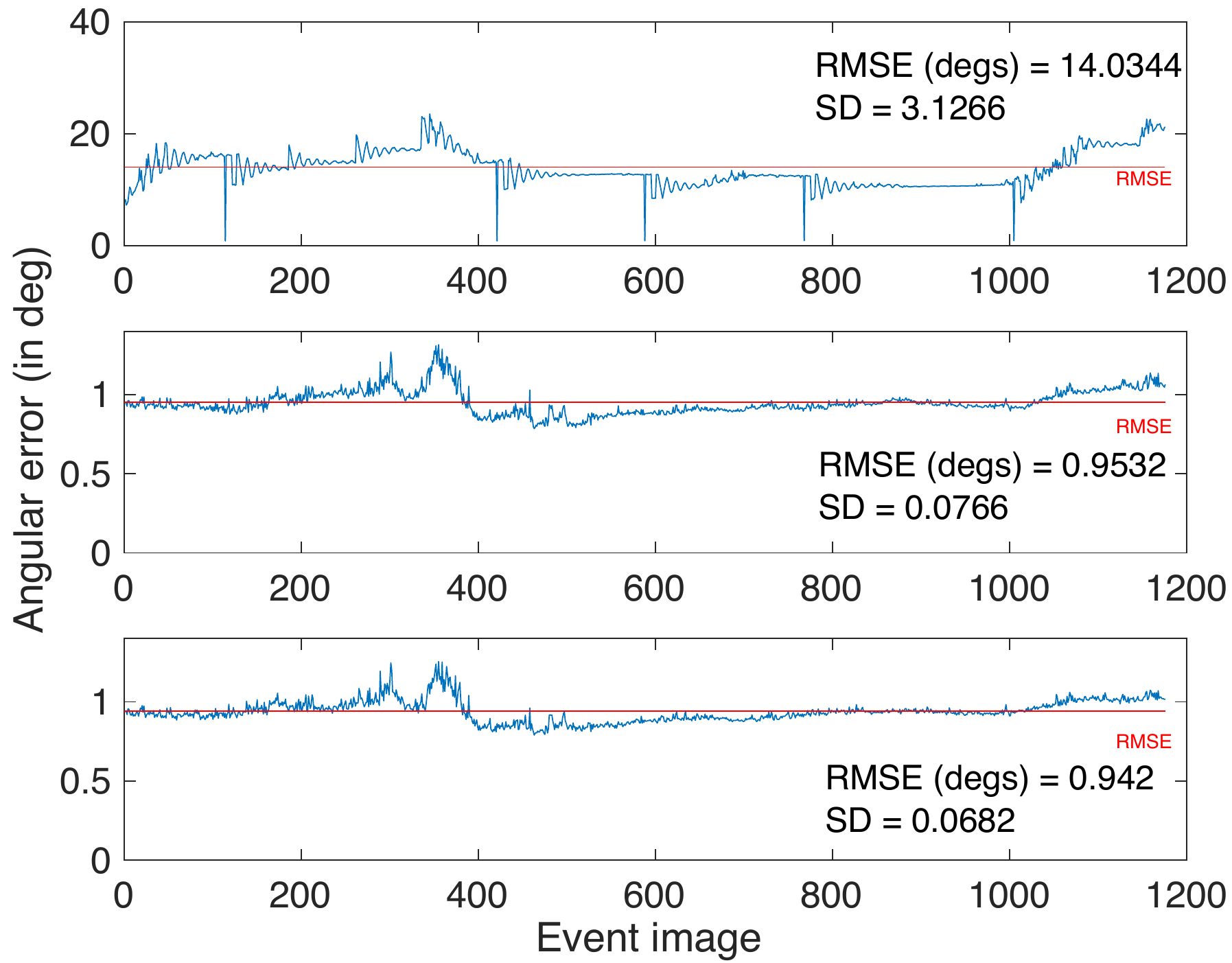}\label{fig:vancouverRABA}}
\caption{Quantitative results on six sequences (more results in supplementary material).}
\label{fig:sixseqs}
\end{figure*}

\section{Conclusion}\label{sec:conclusion}

In this paper, we investigated event cameras for star tracking. The main components in our processing pipeline are novel formulations of rotation averaging and bundle adjustment. We also developed a simulation technique for testing our method. Our results suggest that star tracking using event cameras is feasible and promising.

\clearpage

%------------------------------------------------------------------------

{\small
\bibliographystyle{ieee}
\bibliography{startracking}
}

\end{document}